\documentclass{article}

    \PassOptionsToPackage{numbers, compress}{natbib}

    \usepackage[final]{neurips_2024}

\usepackage[utf8]{inputenc} 
\usepackage[T1]{fontenc}    
\usepackage{hyperref}       
\usepackage{url}            
\usepackage{booktabs}       
\usepackage{amsfonts}       
\usepackage{nicefrac}       
\usepackage{microtype}      
\usepackage{xcolor}         
\usepackage{graphicx}
\usepackage{amsmath}
\usepackage{multirow}
\usepackage{colortbl}

\usepackage{bbding}
\usepackage{fontawesome5}

\title{DiffusionFake: Enhancing Generalization in Deepfake Detection via Guided Stable Diffusion}

\author{
Ke Sun$^{1}$, Shen Chen$^{2}$, Taiping Yao$^{2}$, Hong Liu$^{3}$\thanks{Corresponding Author.},\\
\textbf{Xiaoshuai Sun$^{1}$}, \textbf{Shouhong Ding$^{2}$}, \textbf{Rongrong Ji$^{1}$} \\
$^{1}$ Key Laboratory of Multimedia Trusted Perception and Efficient Computing, \\
Ministry of Education of China, Xiamen University, 361005, P.R. China. \\
$^{2}$ Youtu Lab, Tencent, P.R. China. \\
$^{3}$ Osaka University, Japan. \\
}

\begin{document}

\maketitle

\begin{abstract}

The rapid progress of Deepfake technology has made face swapping highly realistic, raising concerns about the malicious use of fabricated facial content. Existing methods often struggle to generalize to unseen domains due to the diverse nature of facial manipulations. In this paper, we revisit the generation process and identify a universal principle: Deepfake images inherently contain information from both source and target identities, while genuine faces maintain a consistent identity.
Building upon this insight, we introduce DiffusionFake, a novel plug-and-play framework that reverses the generative process of face forgeries to enhance the generalization of detection models. DiffusionFake achieves this by injecting the features extracted by the detection model into a frozen pre-trained Stable Diffusion model, compelling it to reconstruct the corresponding target and source images. This guided reconstruction process constrains the detection network to capture the source and target related features to facilitate the reconstruction, thereby learning rich and disentangled representations that are more resilient to unseen forgeries.
Extensive experiments demonstrate that DiffusionFake significantly improves cross-domain generalization of various detector architectures without introducing additional parameters during inference. Our Codes are available in \href{https://github.com/skJack/DiffusionFake.git}{https://github.com/skJack/DiffusionFake.git}.

\end{abstract}
\section{Introduction}

The rapid progress in AI-generated content (AIGC) has led to the emergence of highly sophisticated forged face content, making it increasingly challenging for humans to distinguish between genuine and forged faces~\cite{rossler2019faceforensics++,zi2020wilddeepfake,dolhansky2020deepfake,chen2024diffusionface}. Face swapping, also known as \textit{Deepfakes}, is one of the most well-known techniques for generating forged facial images. It replaces the face of a target individual with that of a source person to create a seamless and realistic composite image~\cite{tolosana2020deepfakes}. The widespread proliferation of Deepfakes content on social media platforms has raised significant security concerns, including the spread of disinformation, fraud, and impersonation. As a result, developing effective and generalizable face forgery detection methods to counter these malicious attacks has become a critical challenge in the field of computer vision.

The growing diversity of facial forgery techniques has spurred interest in the general face forgery detection task~\cite{sun2021domain,song2022adaptive,luo2023beyond}, which aims to develop models that detect forgeries from unseen domains. 
Previous approaches primarily utilize forgery simulation~\cite{li2020face,shiohara2022detecting,chen2022self,sun2023towards} to augment data by simulating various forgery traces, or framework engineering to enhance generalization through specialized designs like contrastive learning, attention mechanisms, and reconstruction learning~\cite{sun2021dual,zhao2021multi,sun2022information,cao2022end,gu2022hierarchical}. However, their generalization capabilities remain limited due to the reliance on simulating specific forgery artifacts or designing specialized architectures tailored to certain manipulation techniques.

\begin{figure*}[t!]
    \begin{center}
    \includegraphics[width=1.0\textwidth]{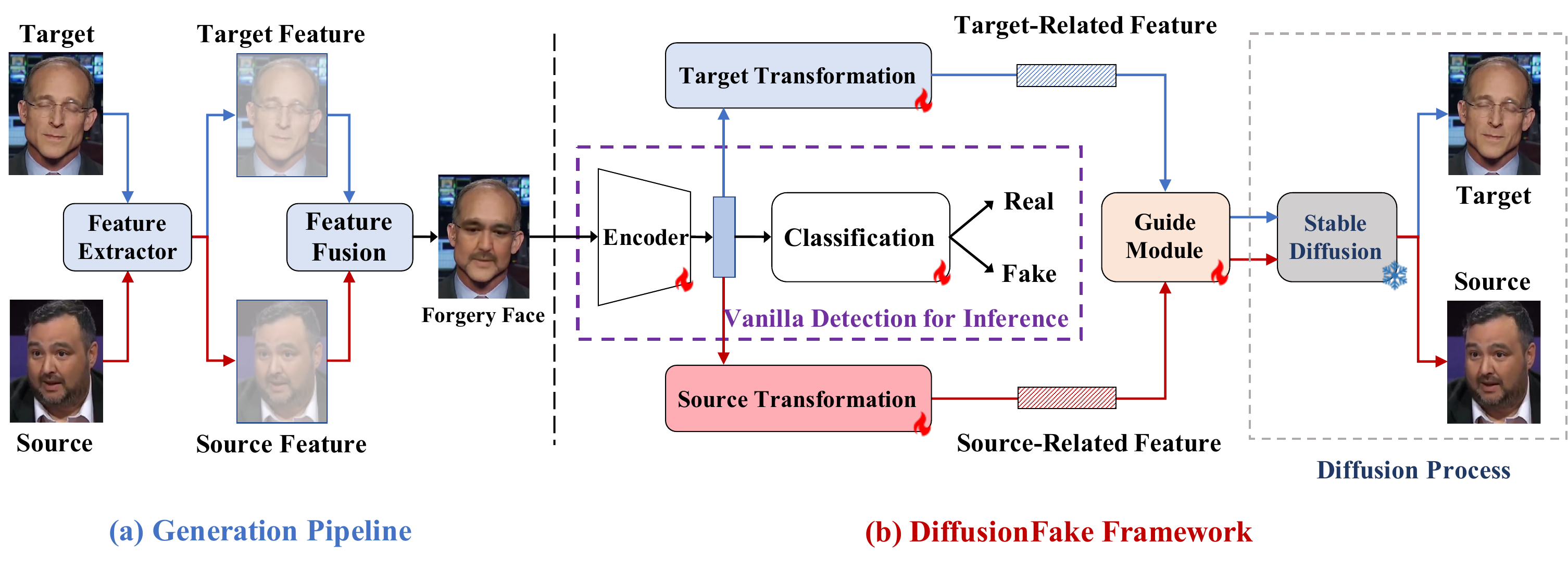}
    \end{center}
    \vspace{-1em}
       \caption{Pipeline of the generation process of Deepfake (a) and our proposed DiffusionFake (b).
       }\vspace{-1em}
    \label{fig:intro}
\end{figure*}

In this paper, we aim to identify the universal features common to all Deepfake faces by revisiting the generative process underlying forged face images. As depicted in Figure~\ref{fig:intro} (a), this process can be distilled into two key steps: (1) a feature extractor module captures salient features from both the source and target images; (2) these features are seamlessly fused through a generalized feature blending module to synthesize a novel Deepfake image. While the specific implementation of feature extraction and fusion may vary across different forgery methods, ranging from learning-based to graphics-based approaches, they all adhere to this fundamental generative paradigm.

Through this analysis, we uncover a crucial insight: {Deepfake images inherently amalgamate information from both source and target faces, whereas genuine images maintain a consistent identity throughout}. This amalgamated information can manifest as low-level artifacts, such as injection noise patterns and spectral discrepancies, or as high-level attributes, including facial expressions and mouth movements, depending on the specific forgery method employed.

Building upon this insight, we raise a question: 
{Can we invert the generative process to extract and leverage the amalgamated source and target features, thereby enhancing the generalization capability of existing forgery detectors?}

To answer this question, we introduce {DiffusionFake}, a novel plug-and-play framework that harnesses the power of Stable Diffusion to guide the forgery detector in learning disentangled source and target features inherent in Deepfakes. The core idea behind DiffusionFake is to inject the features extracted by the detector into a frozen pre-trained Stable Diffusion model, compelling the detector to capture the amalgamated source and target information by optimizing the features to reconstruct the corresponding source and target images.

As illustrated in Figure~\ref{fig:intro} (b), DiffusionFake is a plug-and-play framework that can be seamlessly integrated into existing forgery detectors. The features extracted by the encoder are first passed through the Target and Source Transformation modules, which filter and weight the features to obtain target and source-related representations. These features are then injected into the Stable Diffusion model using a Guide Module, leveraging its pre-trained knowledge to reconstruct the corresponding source and target images and optimize the feature representation. 

During inference, only the encoder and classification modules are used, ensuring no additional parameters or computational overhead. 
By compelling the detector to learn more discriminative and generalized features, DiffusionFake enhances its ability to handle unseen forgeries without compromising efficiency. For example, 
when integrated with EfficientNet-B4, DiffusionFake improves AUC scores on unseen Celeb-DF dataset by around 10\%, demonstrating its effectiveness in enhancing the generalization capability of existing detectors.

The main contributions of our work can be summarized as follows:
\vspace{-1em}
\begin{itemize}
    \item We analyze Deepfake images from a generative perspective and propose a framework that leverages the reverse generation process to enhance the generalization capabilities of face forgery detectors.
    \item We introduce the DiffusionFake framework, a plug-and-play model that integrates a frozen pre-trained Stable Diffusion network to guide the forgery detector in learning disentangled source and target features inherent in Deepfakes, further enhancing the generalization.
    \item Extensive experimental validations demonstrate that the DiffusionFake framework significantly improves generalization capabilities across various architectures without introducing additional inference parameters.
\end{itemize}

\section{Related Work}
\subsection{General Face Forgery Detection}
General face forgery detection aims to improve the performance of forgery detectors on unseen domains and become one of the most critical issues in this field.
Previous work to enhance generalization can be broadly divided into two categories: forgery simulation and framework engineering. The former utilizes data augmentation methods to simulate certain forgery traces, such as blending artifacts~\cite{guillaro2023trufor,li2020face,shiohara2022detecting}, Inconsistency between internal and external faces~\cite{zhao2021learning}, subtle jitter and blur traces~\cite{larue2023seeable}, and fine-grained facial disharmony~\cite{chen2022self,sun2023towards}. The latter improves network architectures or training procedures to help capture more generalized traces.
Such methods approach the problem from different angles. Some employ attention mechanisms to enhance the capture of forgery traces~\cite{zhao2021multi,sun2022information,xu2023tall,shao2023detecting}, while others improve generalization by jointly modeling frequency and spatial domains~\cite{qian2020thinking,li2021frequency,luo2021generalizing,liu2021spatial}. Reconstruction-based methods enhance discriminability against unseen domain forgeries via modeling genuine faces~\cite{chen2021magdr,cao2022end,shi2023real}. Additionally, some approaches use implicit identity as a clue to improve the generalization of Deepfake faces~\cite{huang2023implicit,dong2023implicit} and some explore the local and global relationships of unseen forgeries\cite{chen2021local,bai2023aunet,wang2023dynamic,guillaro2023trufor,dong2022explaining,luo2021capsule}. Furthermore, decoupling methods\cite{liang2022exploring,shao2022detecting,guo2023controllable,le2023quality}, such as ICT~\cite{dong2022protecting} and UCF~\cite{yan2023ucf}, aim to enhance generalization by disentangling different facial information. Our DiffusionFake method addresses this by reversing the forgery process and leveraging pre-trained generative models to complete missing information, enhancing the capture of source-related and target-related features.

\subsection{Diffusion Model}

Diffusion models have emerged as a powerful framework for image generation and manipulation. The seminal work on Denoising Diffusion Probabilistic Models (DDPM)~\cite{ho2020denoising} introduced a novel approach to learn the data distribution by iteratively denoising a Gaussian noise signal. This process allows for high-quality image generation but requires a large number of sampling steps. To address this issue, the Denoising Diffusion Implicit Models (DDIM)~\cite{song2020denoising} proposed a deterministic sampling process that significantly accelerates the generation process while maintaining image quality. Building upon these advancements, the Latent Diffusion Model (LDM)~\cite{rombach2022high} combines the strengths of Variational Autoencoders (VAEs)~\cite{kingma2013auto} and diffusion models. 
By applying the diffusion process in the latent space learned by a VAE, LDM substantially reduces the computational cost and memory requirements during training. 
This innovative architecture has given rise to powerful AIGC pre-trained generative models, such as Stable Diffusion~\footnote{https://stability.ai/news/stable-diffusion-public-release}, which enable high-quality image generation and manipulation with unprecedented efficiency.
Recent developments in controllable diffusion models have further expanded their applicability. ControlNet~\cite{zhang2023adding}introduces a mechanism to guide the image generation process by conditioning the diffusion model on additional control signals, such as segmentation masks or edge maps. Inspired by ControlNet, our DiffusionFake leverages a guide module to inject the source and target-related features into Stable Diffusion to reconstruct the corresponding images.

\section{Methodology}

\begin{figure*}[t!]
    \begin{center}
    \includegraphics[width=1.0\textwidth]{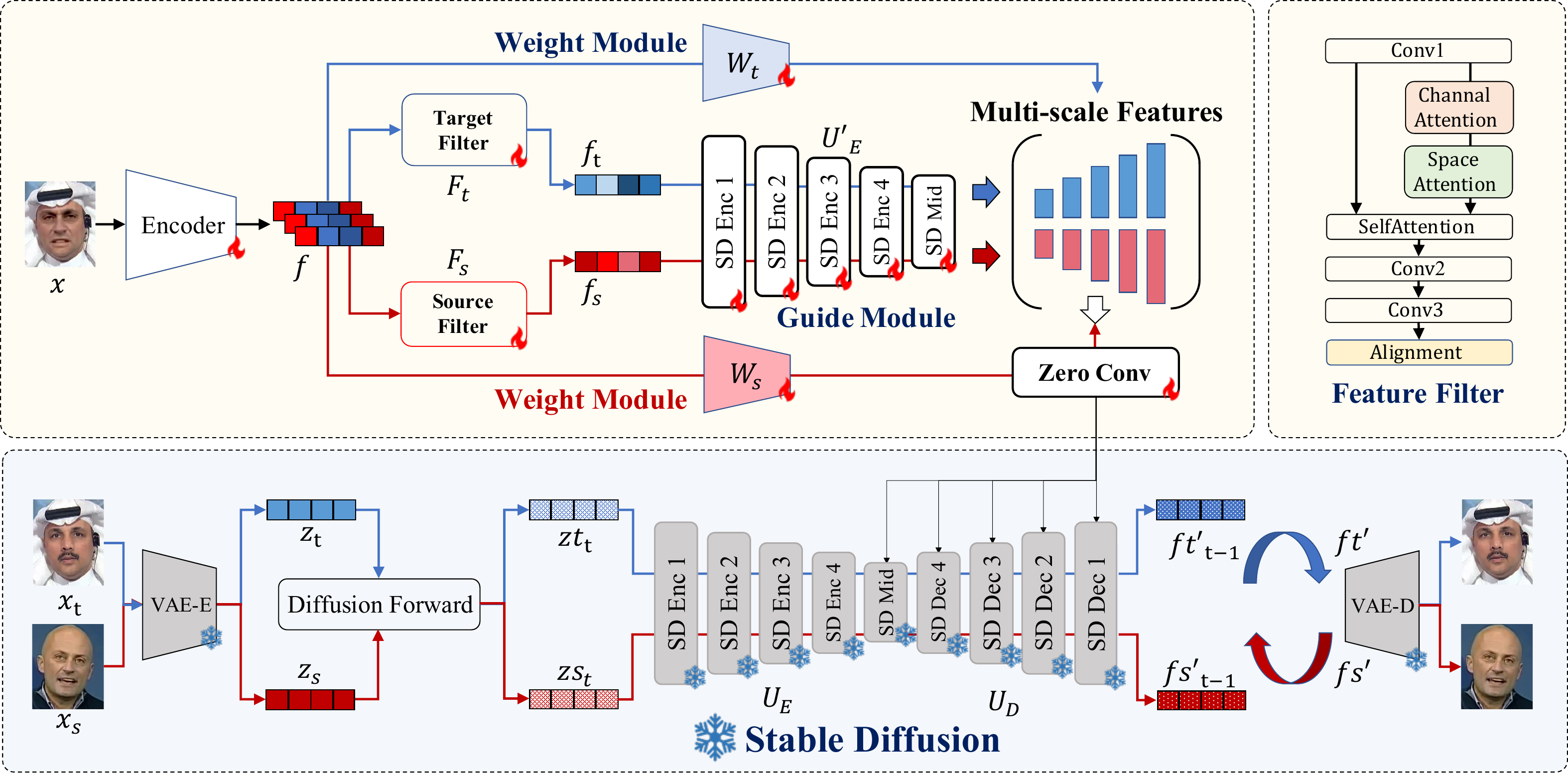}
    \end{center}
    \vspace{-1em}
       \caption{The details of the DiffusionFake method. The blue arrow represents the target branch, the red arrow represents the source branch, the {\footnotesize\color{blue} \SnowflakeChevronBold}~represents the parameter frozen and does not participate in training, and the {\footnotesize\color{red} \faFire} represents the trainable module.
       }
    \label{fig:main}
\end{figure*}

Figure~\ref{fig:main} illustrates the detailed framework of our proposed DiffusionFake method, which aims to enhance the generalization capability of forgery detectors by guiding the learning of amalgamated source and target features through a frozen pre-trained Stable Diffusion model. Specifically, the features extracted by the encoder are first filtered and weighted by the Feature Filter and Weight Modules to obtain source and target-related representations. These features are then injected into a frozen pre-trained Stable Diffusion model via the Guide Module, which reconstructs the corresponding source and target images, compelling the encoder to learn rich and discriminative features.

\subsection{Preliminaries}
\textbf{Diffusion Process.} Denoising Diffusion Probabilistic Models (DDPMs)~\cite{ho2020denoising} are latent variable models that learn to generate data by reversing a gradual noising process. The forward diffusion process gradually adds Gaussian noise to the data $x_0$ according to a variance schedule $\beta_1, \dots, \beta_T$, producing a sequence of noisy samples $x_1, \dots, x_T$. The forward process can be described as:
\begin{equation}
q(x_t|x_{t-1}) = \mathcal{N}(x_t; \sqrt{1-\beta_t}x_{t-1}, \beta_t\mathbf{I})
\end{equation}
The reverse denoising process learns to generate samples from the data distribution by starting with a Gaussian noise sample $x_T$ and iteratively denoising it using a learned denoising function $\epsilon_\theta$. The reverse process is defined as:
\begin{equation}
p_\theta(x_{t-1}|x_t) = \mathcal{N}(x_{t-1}; \mu_\theta(x_t, t), \sigma_t^2\mathbf{I}),
\end{equation}
\begin{equation}
\mu_\theta(x_t, t) = \frac{1}{\sqrt{\alpha_t}} \left( x_t - \frac{1-\alpha_t}{\sqrt{1-\bar{\alpha}t}} \epsilon_\theta(x_t, t) \right),
\end{equation}

$\alpha_t = 1 - \beta_t$, $\bar{\alpha}_t = \prod_{s=1}^t \alpha_s$, and $\sigma_t^2 = \frac{1-\bar{\alpha}_{t-1}}{1-\bar{\alpha}_t}\beta_t$.
The training objective is to minimize the weighted sum of the denoising error at each step:
\begin{equation}
L = \mathbb{E}_{\epsilon \sim \mathcal{N}(0,1), t \sim [1, T]} \left[ || \epsilon - \epsilon_\theta(x_t, t) ||_2^2 \right]
\end{equation}

\textbf{Stable Diffusion.}
The Stable Diffusion Model is a powerful pre-trained model with impressive generative capabilities, able to synthesize various types of images, including different types of human faces.
Built upon the DDPM framework, the Stable Diffusion models employs a Latent Diffusion Model (LDM)~\cite{rombach2022high} to reduce resource consumption. LDM applies the diffusion process in a learned latent space instead of pixel space, which is obtained by training an autoencoder. 
The training objective is:
\begin{equation}
L = \mathbb{E}_{\epsilon \sim \mathcal{N}(0,1), t \sim [1, T]} \left[ || \epsilon - \epsilon_\theta(z_t, t) ||_2^2 \right],
\end{equation}
where $z_t$ is the latent representation encoded by the VAE encoder. This strategic application of latent space modeling not only enhances efficiency but also preserves the high quality of generated images.

\subsection{Feature Transformation}

Given an input image $x$ and its corresponding label $y$, where $y=0$ represents a real face and $y=1$ represents a forged face, let $x_s$ and $x_t$ denote the corresponding source and target images from the training dataset, respectively. For real faces, $x_s$ and $x_t$ are identical to $x$. Let $E$ be the encoder, and the extracted features be $f = E(x)$.
To transform the extracted feature $f$ 
into components that can guide the Stable Diffusion process, we first 
introduce two key modules: the \textit{Feature Filter Module} and the \textit{Weight Module}.

\textbf{Feature Filter Module.} The Feature Filter Module $F$ is to extract source-related and target-related features from the encoded features $f$. To achieve this, we employ two filter networks, $F_s$ and $F_t$, to obtain the source-related feature $f_s = F_s(f)$ and target-related feature $f_t = F_t(f)$, respectively.

As shown in Figure~\ref{fig:main}, the Feature Filter Module combines convolutional layers and attention mechanisms. The features first pass through a convolutional layer to transform the channels. Then channel-wise~\cite{hu2018squeeze} and spatial-wise attention~\cite{woo2018cbam} are applied to adaptively weight and filter the features. These attention mechanisms help to emphasize the most relevant features while suppressing less informative ones, leading to more discriminative representations.

Subsequently, a Multi-Head Attention mechanism~\cite{vaswani2017attention} is then applied to perform cross-attention between the original features (query) and the attention-filtered features (key and value). This operation captures long-range dependencies and enhances the receptive field, enabling more effective feature refinement. 
Finally, to ensure compatibility with the encoder of the Stable Diffusion model, we apply upsampling and pooling operations to align the feature dimensions.


\textbf{Weight Module.} The Weight Module $W$ addresses the varying levels of source and target information embedded in different types of forged images. For example, Deepfakes may evenly blend source and target features, while expression-driven methods like NeuralTextures may predominantly feature target image information with minimal source information confined to specific regions like mouth movements. Uniformly feeding these into the guide module would be suboptimal.

To mitigate this, we use two separate weight modules, $W_s$ and $W_t$, to estimate the information content for source and target features. Each weight module start with a pooling layer, following five MLP layers, and a sigmoid function finally outputs the weight. We train these modules using the similarity scores between the input image and its respective source and target images as ground truth.

Specifically, we encode $x$, $x_t$, and $x_s$ using the pre-trained VAE-Encoder from Stable Diffusion to obtain latent representations $z$, $z_t$, and $z_s$. Such a well-pretrained model can effectively capture and quantify the differences between images, providing a reliable basis for measuring the similarity between the input image and its corresponding source and target images.
The similarity scores between $z$ and $z_t$, and between $z$ and $z_s$, are computed and used to train the weight modules with mean squared error (MSE) loss as follows:
\begin{equation}
\mathcal{L}_{ws} = \left||W_s(f) - \text{sim}(z, z_s)\right||_2^2
\end{equation}
\begin{equation}
\mathcal{L}_{wt} = \left||W_t(f) - \text{sim}(z, z_t)\right||_2^2
\end{equation}
where $\text{sim}(a, b) = \frac{a \cdot b}{|a| |b|}$ denotes the cosine similarity between vectors $a$ and $b$.

By dynamically adjusting the influence of source and target features during the diffusion process, the Weight Module ensures optimal guidance for the Stable Diffusion model, thereby enhancing the encoder's ability to extract generalized features suitable for detecting a wide range of forgeries.

\subsection{Guide Module}

The Guide Module is designed to inject the source-related and target-related features into the frozen pre-trained Stable Diffusion model to guide the reconstruction of the source and target images. As illustrated in Figure~\ref{fig:main}, the Guide Module employs trainable copy and zero convolution layers for feature injection, inspired by ControlNet~\cite{zhang2023adding}.

Let $U_E(\cdot)$ and $U_D(\cdot)$ denote the neural block of the encoder and decoder in the U-Net $\epsilon_\theta$ network of the Stable Diffusion model, respectively. The Guide Module first creates a trainable copy of $U_E(\cdot)$, denoted as $U_E^{'}(\cdot)$. The source-related feature $f_s$ and target-related feature $f_t$ are then independently fed into $U_E^{'}(;)$. The resulting features are combined with the corresponding features from the locked model's decoder $U_D(;)$ using zero convolution layers $Z(\cdot)$, which are $1\times1$ convolutional layers with weights and biases initialized to zeros. This initialization minimizes the impact on the pre-trained model at the beginning of training, stabilizing the training process~\cite{zhang2023adding}. The final output of the Guide Module can be summarized as:



\begin{equation}
fs^{'} = U_D(U_E(z_s)) + Z(U_E^{'}(f_s)) \times W_s(f)
\end{equation}
\begin{equation}
ft^{'} = U_D(U_E(z_t)) + Z(U_E^{'}(f_t)) \times W_t(f)
\end{equation}
where $z_s$ and $z_t$ are the latent representations of the source and target images, respectively, obtained from the pre-trained VAE-Encoder of Stable Diffusion, and $W_s(f)$ and $W_t(f)$ are the weights computed by the Weight Module.

Unlike ControlNet, which aims to control the generated results of the diffusion model, our objective is to optimize the features $f$ by fixing the output and encouraging the capture of more generalizable and disentangled features. By guiding the reconstruction of the source and target images using the respective features, the Guide Module facilitates the learning of rich and discriminative representations that enhance the performance of the forgery detector across various domains and attack types.



During training, we follow a process similar to the LDM~\cite{rombach2022high}, gradually executing the diffusion process, including the time step $t$, to guide the reconstruction of the source and target images. At each time step $t$, the model learns to predict the noise $\epsilon$ that was added to the latent representation of the source or target image. The overall learning objective for the source and target diffusion models can be formulated as:

\begin{equation}
L_s = \mathbb{E}_{\epsilon \sim \mathcal{N}(0,1), t \sim [1, T]} \left[ || \epsilon - \epsilon_\theta(fs_{t}^{'}, t) ||_2^2 \right]
\end{equation}
\begin{equation}
L_t = \mathbb{E}_{\epsilon \sim \mathcal{N}(0,1), t \sim [1, T]} \left[ || \epsilon - \epsilon_\theta(ft_{t}^{'}, t) ||_2^2 \right]
\end{equation}
where ${fs}_{t}^{'}$ and $ft_{t}^{'}$ represent the features within the embedding of time step $t$. 

\subsection{Loss Function and Inference}
We apply a simple binary classification head to the extracted feature $f$ to obtain the predicted label $y'$, which is calculated via typical cross-entropy loss as follows: 
\begin{equation}
L_{ce} = - \left[ y \log y' + (1 - y) \log (1 - y') \right] \label{eq:ce}
\end{equation}

Thus, the final loss function combines Eq.\ref{eq:ce} with the losses from our \textit{Weight module} and \textit{Gudie module}, which is defined as follows:
\begin{equation}
L = L_{ce} + \lambda_s L_s + \lambda_t L_t + L_{ws} + L_{wt}
\end{equation}
where $\lambda_s$ and $\lambda_t$ are hyperparameters that balance the contributions of the source and target diffusion losses, $L_s$ and $L_t$, respectively.

\textbf{Inference.}
During inference, only the Encoder and the classification head are retained, as shown in the purple dotted box in Figure \ref{fig:main}. It is worth noting that DiffusionFake ensures the encoder network extracts generalized features only during training. Consequently, our DiffusionFake framework does not introduce any additional parameters during inference, thereby enhancing generalizability and reducing computational overhead.

\section{Experiment}
\label{sec:experiment}
\subsection{Experimental Setting}
\textbf{Dataset.}
To evaluate the generalization ability of DiffusionFake, we conduct experiments on several challenging datasets:
(1) FaceForensics++ (FF++)~\cite{rossler2019faceforensics++}: This widely-used dataset contains 1,000 videos with four manipulation methods: DeepFakes, NeuralTextures, Face2Face, and FaceSwap. The pairwise real and forged data enable the generation of mixed forgery images. (2) Celeb-DF~\cite{li2019celeb}: A high-quality DeepFake dataset containing various scenarios.  (3) DeepFake Detection (DFD): This dataset comprises 363 real videos and 3,068 fake videos, primarily generated using the DeepFake method. (4) DFDC Preview (DFDC-P)~\cite{dolhansky2020deepfake}: A challenging dataset with 1,133 real videos and 4,080 fake videos, featuring various manipulation methods and backgrounds. (5) WildDeepfake~\cite{zi2020wilddeepfake}: A diverse dataset obtained from the internet, capturing a wide range of real-world scenarios. (6) DiffSwap~\cite{chen2024diffusionface}: A recently released dataset containing 30,000 high-quality face swaps generated using the diffusion-based DiffSwap method~\cite{zhao2023diffswap} on the MM-Celeb-A dataset. This dataset allows for evaluating cross-method generalization.

\textbf{Training details.}
DiffusionFake is a plug-and-play architecture that can be integrated with different backbone networks by simply adjusting the dimensions of the alignment layer in the Feature Filter module. During training, we utilize a pre-trained Stable Diffusion 1.5 model with frozen parameters. Input images are resized to 224$\times$224 pixels. We employ the Adam optimizer with a learning rate of 1e-5 and a batch size of 32. The model is trained for 20 epochs. The hyperparameters $\lambda_s$ and $\lambda_t$ are set to 0.7 and 1, respectively. We employ widely used data augmentations, such as HorizontalFlip, and CutOut. To ensure a fair comparison, we follow the data split strategy used in FaceForensics++~\cite{rossler2019faceforensics++}. The overall framework is implemented in Pytorch on one NVIDIA A-100 GPU.

\begin{table*}[!t]
\renewcommand\arraystretch{1.2}
\centering
\caption{{Frame-level} cross-database evaluation from FF++(HQ) to Celeb-DF, Wild Deepfake, DFDC-P, DFD, and DiffSwap in terms of AUC and EER. * represents the results reproduced using open-source code or model.
}\vspace{0.5em}
\label{table:1}
\resizebox{\textwidth}{!}{
\begin{tabular}{l|cccccccccc|ccc}
\hline
\multirow{2}*{Method} & \multicolumn{2}{c}{Celeb-DF} & \multicolumn{2}{c}{Wild Deepfake} & \multicolumn{2}{c}{DFDC-P} & \multicolumn{2}{c}{DFD} & \multicolumn{2}{c}{DiffSwap} &
\multicolumn{2}{c}{Average}
\\
\cmidrule{2-13}
& AUC & EER & AUC & EER & AUC & EER & AUC & EER & AUC & EER & AUC & EER\\
\hline
Xception~\cite{chollet2017xception} & 65.27 & 38.77 & 66.17 & 40.14 & 69.80 & 35.41 & 87.86 & 21.04 & 74.25 & 32.04
& 72.67 & 33.48 \\
Face X-ray~\cite{tan2019efficientnet} & 74.20 & - & - & - & 70.00 & - & 85.60 & - & - & -
& - & -\\
F3-Net*~\cite{qian2020thinking} & 71.21 & 34.03 & 67.71 & 40.17 & 72.88 & 33.38 & 86.10 & 26.17 & 76.89 & 30.83 & 74.96 & 32.92\\
MAT*~\cite{zhao2021multi} & 70.65 & 35.83 & 70.15 & 36.53 & 67.34 & 38.31 & 87.58 & 21.73 & 79.93 & 27.77
& 75.13 & 32.03\\
GFF*~\cite{luo2021generalizing} & 75.31 & 32.48 & 66.51 & 41.52 & 71.58 & 34.77 & 85.51 & 25.64 & 78.38 & 28.15
& 75.46 & 32.51\\
LTW~\cite{sun2021domain} & 77.14 & 29.34 & 67.12 & 39.22 & 74.58 & 33.81 & 88.56 & 20.57 & 77.95 & 29.01
& 77.07 & 30.39\\
LRL~\cite{chen2021local} & 78.26 & 29.67 & 68.76 & 37.50 & 76.53 & 32.41 & 89.24 & 20.32 &- & -
&- & -\\
DCL~\cite{sun2021dual} & 82.30 & 26.53 & 71.14 & 36.17 & 76.71 & 31.97 & 91.66 & 16.63 & 80.21 & 27.37
& 80.40& 27.73\\
PCL+I2G~\cite{zhao2021learning} & 81.80 & - & - & - & - & - & - & - & - & -
& - & -\\
SBI*~\cite{shiohara2022detecting} & 80.76 & 26.97 & 68.22 & 38.11 & 76.53 & 30.22 & 88.13 & 17.25 & 75.20 & 31.49
& 77.77 & 28.81\\
UIA-ViT~\cite{zhuang2022uia} & 82.41 & - & - & - & 75.80 & - & \textbf{94.68} & - & - & -
& - & -\\
RECCE*~\cite{cao2022end} & 70.50 & 35.34 & 67.93 & 39.82 & 75.88 & 32.41 & 89.91 & 19.95 & 77.59 & 29.38
& 76.36 & 31.38\\
UCF~\cite{yan2023ucf} & 75.27 & - & - & - & 75.94 & - & 80.74 & - & - & -
& - & -\\
CADDM*~\cite{dong2023implicit} & 77.56 & 30.63 & 72.56 & 33.63 & 72.45 & 33.56 & 82.90 & 25.20 & 75.58 & 31.01
& 76.21 & 30.81\\
EN-b4*~\cite{tan2019efficientnet} & 73.51 & 34.17 & 70.04 & 37.03 & 70.51 & 33.98 & 87.57 & 21.31 & 77.38 & 29.44
& 75.80 & 31.19\\
VIT-B*~\cite{tan2019efficientnet} & 74.64 & 33.07 & 75.46 & 31.53 & 74.24 & 34.29 & 84.38 & 24.15 & 78.50 & 28.14
& 77.44& 30.24\\
\hline
En-b4+Ours & \textbf{83.17} & \textbf{24.59} & 75.17 & 33.25 & 77.35 & 30.17 & 91.71 & \textbf{16.27} & 82.02 & 25.55
& 81.88& 25.97\\
VIT-B+Ours & 80.46 & 27.51 & \textbf{80.14} & \textbf{29.62} & \textbf{80.95} & \textbf{27.66} & 90.36 &19.73 & \textbf{86.98} & \textbf{21.32}
& \textbf{83.78}& \textbf{25.17}\\
\hline
\end{tabular}\vspace{-1.5em}
}
\end{table*}
\subsection{Experimental Results}
We use AUC and EER to evaluate all the methods, including ours, both of them are widely used in deepfake detection.\footnote{More details about the evaluation metrics can be found in the appendix.}
We compare DiffusionFake with several state-of-the-art methods.

\textbf{Cross-dataset evaluation.} To validate the generalization capability of DiffusionFake, we first evaluate its performance on unseen datasets against recent state-of-the-art methods. Following previous settings, we train the models on the FF++ dataset and test them on several unseen domain datasets. The frame-level results are shown in Table~\ref{table:1}, where * denotes results obtained using official code with consistent training settings and data.

We evaluate its performance using two representative backbones: EfficientNet-B4 (En-B4) and ViT-B. We observe that incorporating DiffusionFake significantly improves the generalization ability of both architectures compared to their original classification backends. For En-B4, our method boosts performance on Celeb-DF by 11\% and achieves an average improvement of 6\%. Similarly, ViT-B sees a 6\% increase on DFDC and an average gain of 6\% when trained with DiffusionFake. Remarkably, \textbf{these enhancements are achieved without increasing the parameter count or computational overhead during inference}.
Compared to state-of-the-art methods, DiffusionFake outperforms disentanglement-based approaches like UCF and CAADM on Celeb-DF. Moreover, our method demonstrates substantial improvements on the latest diffusion-based face swapping dataset, DiffSwap, highlighting its effectiveness against the most recent forgery techniques. These results validate the ability of the guide module and Stable Diffusion network to encourage the encoder to learn more generalizable features by reconstructing source and target images. Due to space limitations, we provide the results of single-source and multi-source cross-manipulation evaluations in the appendix.

\subsection{Ablation Study}

\textbf{Ablation of components.}
We conducted an ablation study to investigate the impact of the key modules in DiffusionFake: 1) the pre-trained Stable Diffusion (SD) model, 2) the Feature Filter module, and 3) the Weight Module. The results are shown in Table~\ref{table:ablation}, where without SD refers to not loading the pre-trained weights of the SD model, and without Filter means directly feeding the encoder's output features $f$ into the guide module.

We can observe that the pre-trained Stable Diffusion model is crucial for the DiffusionFake framework. Without the pre-trained weights, the network struggles to reconstruct the source and target images due to information loss, hindering the training process. Furthermore Both the Feature Filter and Weight modules play significant roles, and removing either of them leads to a performance decline. Specifically, eliminating the Filter module results in a 5\% AUC drop, as the filtering component allows the reconstruction to focus on relevant information without interference from redundant features. On the other hand, the absence of the Weight module causes a 3\% performance decrease, as this module assesses the amount of source and target information contained in the image, providing a prior for the generative network to determine the importance of guided information during the reconstruction process.

\textbf{Ablation of backbones.}
As our method can be flexibly embedded into different backbones by adjusting the alignment of the Feature Filter, we conduct an ablation study on various backbone architectures to demonstrate the versatility of DiffusionFake. We experiment with traditional ResNet-34, lightweight EfficientNet-B0, and the ViT-based ViT-Small. The results in Table~\ref{table:backbone} show that integrating our method into these backbones significantly improves generalization performance. For instance, applying DiffusionFake to the lightweight EfficientNet-B0 increases the generalization accuracy on Celeb-DF from 71.75\% to 76.31\%, surpassing the original EfficientNet-B4 (73.51\%). This evidence suggests that our method can effectively drive different encoders to extract more generalizable features.

\begin{table}[t!]
    \begin{minipage}[t]{0.48\textwidth}
        \centering
        \caption{Abalation study of different components of DiffusionFake.}
        \label{table:ablation}
        \renewcommand\arraystretch{1.2}
        \resizebox{\columnwidth}{!}{
            \begin{tabular}{c|c|c|cc|cc}
        \hline
        \multirow{2}*{SD}&\multirow{2}*{Filter}&\multirow{2}*{Weight} & \multicolumn{2}{c|}{Celeb-DF}&
        \multicolumn{2}{c}{DFDC-P}  \\

        \cline{4-7}
        &&& AUC& EER &AUC& EER \\
        \hline
        $\times$&$\times$&$\times$&71.87&34.28&71.78&35.01\\
        $\times$&$\checkmark$&$\checkmark$&73.87&32.06&72.41&34.25\\
        $\checkmark$&$\times$&$\times$&77.35&29.05&75.69&32.12\\

        $\checkmark$&$\checkmark$&$\times$&80.79&26.37&76.17&31.57\\
        $\checkmark$&$\times$&$\checkmark$&78.67&28.33&76.59&31.22\\
        $\checkmark$&$\checkmark$&$\checkmark$&83.17&24.59&77.35&30.17\\
        

        \hline
    \end{tabular}
        }

        \end{minipage}
        \begin{minipage}[t]{0.48\textwidth}
        \centering
        \caption{Abalation study of backbones.}
        \label{table:backbone}
        \renewcommand\arraystretch{1.2}
        \resizebox{\columnwidth}{!}{
        \begin{tabular}{c|cc|cc}
        \hline
        \multirow{2}*{Backbone} & \multicolumn{2}{c|}{Celeb-DF}&
        \multicolumn{2}{c}{WDF}  \\

        \cline{2-5}

        & AUC& EER &AUC& EER \\
        \hline
        ResNet&68.89&36.78&69.91&38.07\\
        ResNet+Ours&75.27&32.44&73.25&34.27\\
        \hline
        En-b0&71.74&34.56&69.24&38.32\\
        En-b0+Ours&76.31&31.56&74.40&33.99\\
        \hline
        Vit-S&70.59&35.87&70.60&37.59\\
        Vit-S+Ours&74.58&32.95&75.10&33.87\\

        \hline
    \end{tabular}
        }

        \end{minipage}
    \end{table}

\begin{figure*}[t!]
    \vspace{-1em}
    \begin{center}
    \includegraphics[width=0.95\textwidth]{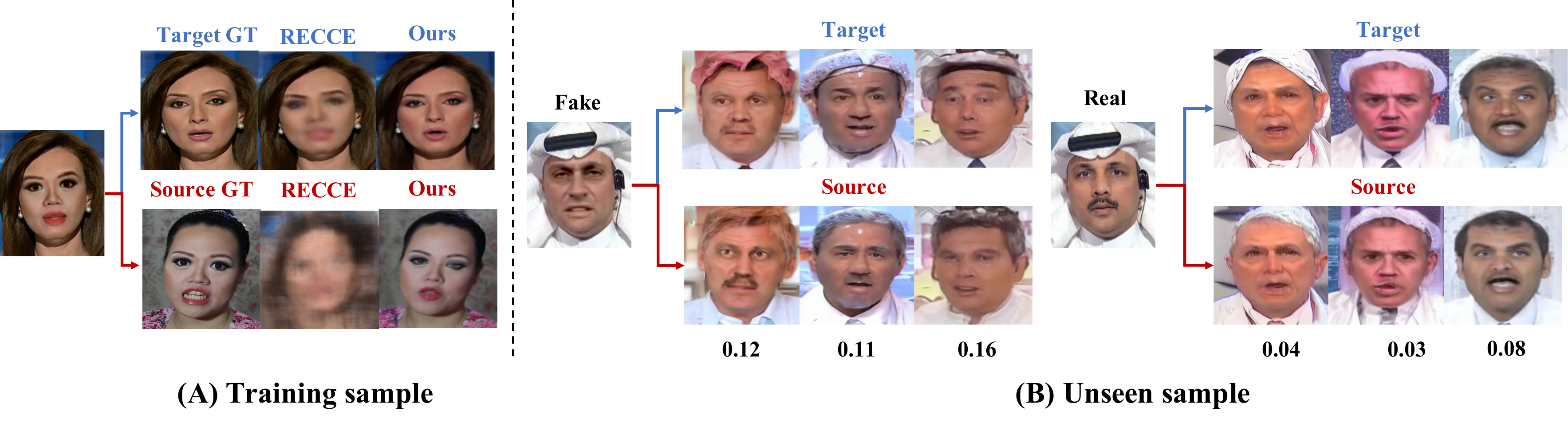}
    \end{center}
    \vspace{-1em}
       \caption{Reconstruction results of DiffusionFake for training (A) and unseen (B) samples. For unseen samples, the model is provided with three sets of initial Gaussian noise, differing only in the injected guide information. The numbers below represent the Euclidean distance between the corresponding source and target features.
       }\vspace{-1em}
    \label{fig:vis_rec}
\end{figure*}

\subsection{Analysis and Visualizations}

\textbf{Visualizations of reverse results.}
Figure \ref{fig:vis_rec} showcases the reconstruction results for both training and unseen samples using DiffusionFake. For training samples (Figure \ref{fig:vis_rec} A), DiffusionFake effectively reconstructs the target image, despite the source image being slightly blurry due to information loss, capturing the basic characteristics of the ground truth. 
In order to compare the reconstruction effect more intuitively, we use the RECCE method to directly reconstruct the source and target images of the fake image. It can be seen that the reconstruction effect is very poor. In contrast, the reconstruction effect of our method is better due to the help of the pre-trained SD model.
For unseen samples (Figure \ref{fig:vis_rec} B), fake images with mixed features result in significant differences between reconstructed target and source images, while real images exhibit smaller differences. The Euclidean distances at the bottom quantify these differences, indicating larger differences in fake images compared to real ones.

\begin{figure*}[t!]
    \begin{center}
    \includegraphics[width=1.0\textwidth]{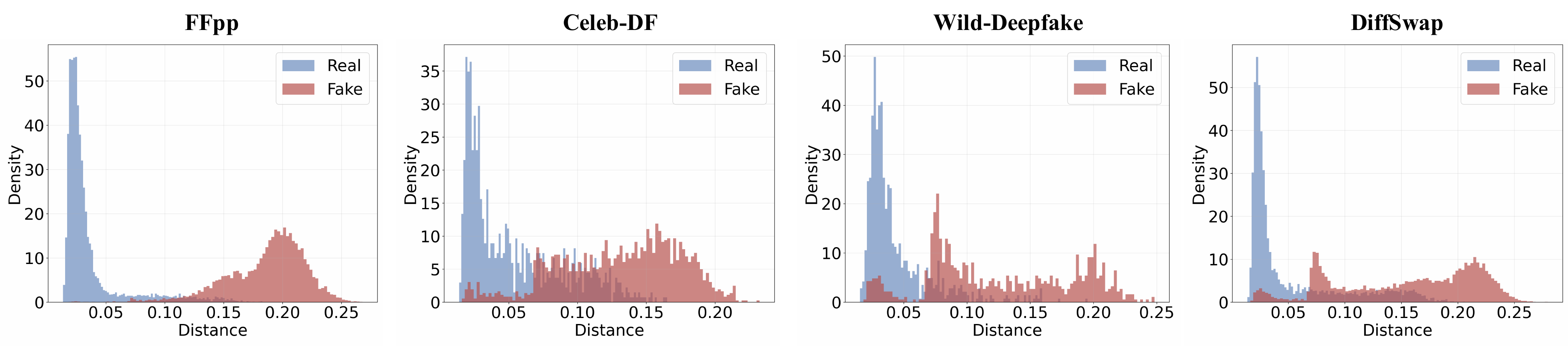}
    \end{center}
    \vspace{-1em}
       \caption{Histogram of feature divergence on FFpp, Celeb-DF, Wild-Deepfake, and DiffSwap.
       }
       
    \label{fig:vis_hist}
\end{figure*}

\textbf{Analysis of feature divergence. }
DiffusionFake utilizes two Feature Filter modules to separate source-related and target-related features, expecting significant divergence between $f_s$ and $f_t$ for forged images and minimal differences for genuine faces. To validate this, we visualize the Euclidean distance distribution between these features across various datasets, including FFpp, Celeb-DF, Wild-Deepfake, and DiffSwap, as shown in Figure \ref{fig:vis_hist}. The plots clearly distinguish real from fake samples: real samples have small feature distances, mostly within 0.05, whereas fake samples show larger distances due to mixed source and target information. These observations strongly support that DiffusionFake effectively disentangles source and target information in the extracted features.


\begin{figure*}[t!]
    \begin{center}
    \includegraphics[width=0.95\textwidth]{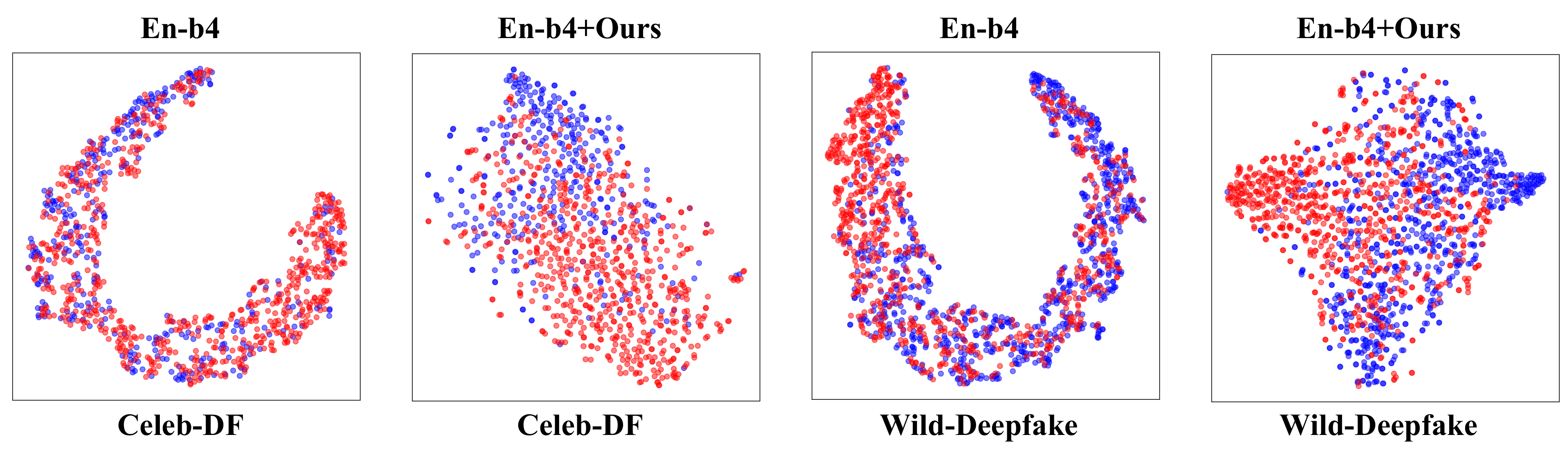}
    \end{center}
    \vspace{-1em}
       \caption{Feature distribution of En-b4 model and the En-b4 model trained with our DiffusionFace on two unseen datasets Celeb-DF and Wild-Deepfake via t-SNE. The red represents the real samples while the blue represents the fake ones.
       }
    \label{fig:tsne}
\end{figure*}

\textbf{Analysis of feature distribution.}
To demonstrate that DiffusionFake enhances the discriminative power and generalization ability of the extracted features, we visualize the t-SNE plots of the last layer features from two encoders: the original EfficientNet-B4 (En-B4) and En-B4 trained with DiffusionFake. The feature distributions are examined on two unseen datasets, Celeb-DF and Wild-Deepfake.
As illustrated in Figure \ref{fig:tsne}, the original En-B4 exhibits poor generalization on both datasets, with the real and fake features being nearly inseparable. In contrast, when trained with DiffusionFake, the encoder learns to capture the generalizable hybrid features present in forged images via the reverse process. Consequently, the real and fake features become more distinctly separated, forming clear decision boundaries on both unseen datasets.

\section{Conclusion}
In this paper, we introduce DiffusionFake, a novel framework that leverages the generative process of face forgery to enhance the generalization capabilities of detection models. DiffusionFake inverts this generative process to extract and utilize hybrid features from source and target identities for effective forgery detection. Extensive experiments demonstrate that DiffusionFake significantly improves the generalization performance of various detector architectures without increasing inference parameters. 
The proposed framework enables the learning of discriminative and generalizable features, enhancing the robustness of detectors against a wide range of unseen forgeries.

{\small
    \bibliographystyle{plain}
    \bibliography{main}

\begin{thebibliography}{10}

\bibitem{bai2023aunet}
Weiming Bai, Yufan Liu, Zhipeng Zhang, Bing Li, and Weiming Hu.
\newblock Aunet: Learning relations between action units for face forgery detection.
\newblock In {\em Proceedings of the IEEE/CVF Conference on Computer Vision and Pattern Recognition}, pages 24709--24719, 2023.

\bibitem{cao2022end}
Junyi Cao, Chao Ma, Taiping Yao, Shen Chen, Shouhong Ding, and Xiaokang Yang.
\newblock End-to-end reconstruction-classification learning for face forgery detection.
\newblock In {\em Proceedings of the IEEE/CVF Conference on Computer Vision and Pattern Recognition}, pages 4113--4122, 2022.

\bibitem{chen2022self}
Liang Chen, Yong Zhang, Yibing Song, Lingqiao Liu, and Jue Wang.
\newblock Self-supervised learning of adversarial example: Towards good generalizations for deepfake detection.
\newblock In {\em CVPR}, pages 18710--18719, 2022.

\bibitem{chen2021local}
Shen Chen, Taiping Yao, Yang Chen, Shouhong Ding, Jilin Li, and Rongrong Ji.
\newblock Local relation learning for face forgery detection.
\newblock {\em AAAI}, 2021.

\bibitem{chen2021magdr}
Zhikai Chen, Lingxi Xie, Shanmin Pang, Yong He, and Bo~Zhang.
\newblock Magdr: Mask-guided detection and reconstruction for defending deepfakes.
\newblock In {\em Proceedings of the IEEE/CVF Conference on Computer Vision and Pattern Recognition}, pages 9014--9023, 2021.

\bibitem{chen2024diffusionface}
Zhongxi Chen, Ke~Sun, Ziyin Zhou, Xianming Lin, Xiaoshuai Sun, Liujuan Cao, and Rongrong Ji.
\newblock Diffusionface: Towards a comprehensive dataset for diffusion-based face forgery analysis.
\newblock {\em arXiv preprint arXiv:2403.18471}, 2024.

\bibitem{chollet2017xception}
Fran{\c{c}}ois Chollet.
\newblock Xception: Deep learning with depthwise separable convolutions.
\newblock In {\em CVPR}, pages 1251--1258, 2017.

\bibitem{dolhansky2020deepfake}
Brian Dolhansky, Joanna Bitton, Ben Pflaum, Jikuo Lu, Russ Howes, Menglin Wang, and Cristian~Canton Ferrer.
\newblock The deepfake detection challenge dataset.
\newblock {\em arXiv preprint arXiv:2006.07397}, 2020.

\bibitem{dong2023implicit}
Shichao Dong, Jin Wang, Renhe Ji, Jiajun Liang, Haoqiang Fan, and Zheng Ge.
\newblock Implicit identity leakage: The stumbling block to improving deepfake detection generalization.
\newblock In {\em Proceedings of the IEEE/CVF Conference on Computer Vision and Pattern Recognition}, pages 3994--4004, 2023.

\bibitem{dong2022explaining}
Shichao Dong, Jin Wang, Jiajun Liang, Haoqiang Fan, and Renhe Ji.
\newblock Explaining deepfake detection by analysing image matching.
\newblock In {\em European Conference on Computer Vision}, pages 18--35. Springer, 2022.

\bibitem{dong2022protecting}
Xiaoyi Dong, Jianmin Bao, Dongdong Chen, Ting Zhang, Weiming Zhang, Nenghai Yu, Dong Chen, Fang Wen, and Baining Guo.
\newblock Protecting celebrities from deepfake with identity consistency transformer.
\newblock In {\em Proceedings of the IEEE/CVF Conference on Computer Vision and Pattern Recognition}, pages 9468--9478, 2022.

\bibitem{gu2022hierarchical}
Zhihao Gu, Taiping Yao, Yang Chen, Shouhong Ding, and Lizhuang Ma.
\newblock Hierarchical contrastive inconsistency learning for deepfake video detection.
\newblock In {\em European Conference on Computer Vision}, pages 596--613. Springer, 2022.

\bibitem{guillaro2023trufor}
Fabrizio Guillaro, Davide Cozzolino, Avneesh Sud, Nicholas Dufour, and Luisa Verdoliva.
\newblock Trufor: Leveraging all-round clues for trustworthy image forgery detection and localization.
\newblock In {\em Proceedings of the IEEE/CVF Conference on Computer Vision and Pattern Recognition}, pages 20606--20615, 2023.

\bibitem{guo2023controllable}
Ying Guo, Cheng Zhen, and Pengfei Yan.
\newblock Controllable guide-space for generalizable face forgery detection.
\newblock In {\em Proceedings of the IEEE/CVF International Conference on Computer Vision}, pages 20818--20827, 2023.

\bibitem{ho2020denoising}
Jonathan Ho, Ajay Jain, and Pieter Abbeel.
\newblock Denoising diffusion probabilistic models.
\newblock {\em Advances in neural information processing systems}, 33:6840--6851, 2020.

\bibitem{hu2018squeeze}
Jie Hu, Li~Shen, and Gang Sun.
\newblock Squeeze-and-excitation networks.
\newblock In {\em CVPR}, pages 7132--7141, 2018.

\bibitem{huang2023implicit}
Baojin Huang, Zhongyuan Wang, Jifan Yang, Jiaxin Ai, Qin Zou, Qian Wang, and Dengpan Ye.
\newblock Implicit identity driven deepfake face swapping detection.
\newblock In {\em Proceedings of the IEEE/CVF Conference on Computer Vision and Pattern Recognition}, pages 4490--4499, 2023.

\bibitem{kingma2013auto}
Diederik~P Kingma and Max Welling.
\newblock Auto-encoding variational bayes.
\newblock {\em International Conference on Learning Representations}, 2014.

\bibitem{larue2023seeable}
Nicolas Larue, Ngoc-Son Vu, Vitomir Struc, Peter Peer, and Vassilis Christophides.
\newblock Seeable: Soft discrepancies and bounded contrastive learning for exposing deepfakes.
\newblock In {\em Proceedings of the IEEE/CVF International Conference on Computer Vision}, pages 21011--21021, 2023.

\bibitem{le2023quality}
Binh~M Le and Simon~S Woo.
\newblock Quality-agnostic deepfake detection with intra-model collaborative learning.
\newblock In {\em Proceedings of the IEEE/CVF International Conference on Computer Vision}, pages 22378--22389, 2023.

\bibitem{li2021frequency}
Jiaming Li, Hongtao Xie, Jiahong Li, Zhongyuan Wang, and Yongdong Zhang.
\newblock Frequency-aware discriminative feature learning supervised by single-center loss for face forgery detection.
\newblock In {\em Proceedings of the IEEE/CVF conference on computer vision and pattern recognition}, pages 6458--6467, 2021.

\bibitem{li2020face}
Lingzhi Li, Jianmin Bao, Ting Zhang, Hao Yang, Dong Chen, Fang Wen, and Baining Guo.
\newblock Face x-ray for more general face forgery detection.
\newblock In {\em CVPR}, pages 5001--5010, 2020.

\bibitem{li2019celeb}
Yuezun Li, Xin Yang, Pu~Sun, Honggang Qi, and Siwei Lyu.
\newblock Celeb-df: A new dataset for deepfake forensics.
\newblock {\em arXiv preprint arXiv:1909.12962}, 2019.

\bibitem{liang2022exploring}
Jiahao Liang, Huafeng Shi, and Weihong Deng.
\newblock Exploring disentangled content information for face forgery detection.
\newblock In {\em European Conference on Computer Vision}, pages 128--145. Springer, 2022.

\bibitem{liu2021spatial}
Honggu Liu, Xiaodan Li, Wenbo Zhou, Yuefeng Chen, Yuan He, Hui Xue, Weiming Zhang, and Nenghai Yu.
\newblock Spatial-phase shallow learning: rethinking face forgery detection in frequency domain.
\newblock In {\em CVPR}, pages 772--781, 2021.

\bibitem{luo2023beyond}
Anwei Luo, Chenqi Kong, Jiwu Huang, Yongjian Hu, Xiangui Kang, and Alex~C Kot.
\newblock Beyond the prior forgery knowledge: Mining critical clues for general face forgery detection.
\newblock {\em IEEE Transactions on Information Forensics and Security}, 19:1168--1182, 2023.

\bibitem{luo2021capsule}
Anwei Luo, Enlei Li, Yongliang Liu, Xiangui Kang, and Z~Jane Wang.
\newblock A capsule network based approach for detection of audio spoofing attacks.
\newblock In {\em ICASSP 2021-2021 IEEE International Conference on Acoustics, Speech and Signal Processing (ICASSP)}, pages 6359--6363. IEEE, 2021.

\bibitem{luo2021generalizing}
Yuchen Luo, Yong Zhang, Junchi Yan, and Wei Liu.
\newblock Generalizing face forgery detection with high-frequency features.
\newblock In {\em CVPR}, pages 16317--16326, 2021.

\bibitem{qian2020thinking}
Yuyang Qian, Guojun Yin, Lu~Sheng, Zixuan Chen, and Jing Shao.
\newblock Thinking in frequency: Face forgery detection by mining frequency-aware clues.
\newblock In {\em ECCV}, pages 86--103. Springer, 2020.

\bibitem{rombach2022high}
Robin Rombach, Andreas Blattmann, Dominik Lorenz, Patrick Esser, and Bj{\"o}rn Ommer.
\newblock High-resolution image synthesis with latent diffusion models.
\newblock In {\em Proceedings of the IEEE/CVF conference on computer vision and pattern recognition}, pages 10684--10695, 2022.

\bibitem{rossler2019faceforensics++}
Andreas Rossler, Davide Cozzolino, Luisa Verdoliva, Christian Riess, Justus Thies, and Matthias Nie{\ss}ner.
\newblock Faceforensics++: Learning to detect manipulated facial images.
\newblock In {\em ICCV}, pages 1--11, 2019.

\bibitem{shao2022detecting}
Rui Shao, Tianxing Wu, and Ziwei Liu.
\newblock Detecting and recovering sequential deepfake manipulation.
\newblock In {\em European Conference on Computer Vision}, pages 712--728. Springer, 2022.

\bibitem{shao2023detecting}
Rui Shao, Tianxing Wu, and Ziwei Liu.
\newblock Detecting and grounding multi-modal media manipulation.
\newblock In {\em Proceedings of the IEEE/CVF Conference on Computer Vision and Pattern Recognition}, pages 6904--6913, 2023.

\bibitem{shi2023real}
Liang Shi, Jie Zhang, and Shiguang Shan.
\newblock Real face foundation representation learning for generalized deepfake detection.
\newblock {\em arXiv preprint arXiv:2303.08439}, 2023.

\bibitem{shiohara2022detecting}
Kaede Shiohara and Toshihiko Yamasaki.
\newblock Detecting deepfakes with self-blended images.
\newblock In {\em CVPR}, pages 18720--18729, 2022.

\bibitem{song2020denoising}
Jiaming Song, Chenlin Meng, and Stefano Ermon.
\newblock Denoising diffusion implicit models.
\newblock In {\em International Conference on Learning Representations}, 2020.

\bibitem{song2022adaptive}
Luchuan Song, Zheng Fang, Xiaodan Li, Xiaoyi Dong, Zhenchao Jin, Yuefeng Chen, and Siwei Lyu.
\newblock Adaptive face forgery detection in cross domain.
\newblock In {\em European Conference on Computer Vision}, pages 467--484. Springer, 2022.

\bibitem{sun2023towards}
Ke~Sun, Shen Chen, Taiping Yao, Xiaoshuai Sun, Shouhong Ding, and Rongrong Ji.
\newblock Towards general visual-linguistic face forgery detection.
\newblock {\em arXiv preprint arXiv:2307.16545}, 2023.

\bibitem{sun2022information}
Ke~Sun, Hong Liu, Taiping Yao, Xiaoshuai Sun, Shen Chen, Shouhong Ding, and Rongrong Ji.
\newblock An information theoretic approach for attention-driven face forgery detection.
\newblock In {\em European Conference on Computer Vision}, pages 111--127. Springer, 2022.

\bibitem{sun2021domain}
Ke~Sun, Hong Liu, Qixiang Ye, Jianzhuang Liu, Yue Gao, Ling Shao, and Rongrong Ji.
\newblock Domain general face forgery detection by learning to weight.
\newblock In {\em AAAI}, volume~35, pages 2638--2646, 2021.

\bibitem{sun2021dual}
Ke~Sun, Taiping Yao, Shen Chen, Shouhong Ding, Rongrong Ji, et~al.
\newblock Dual contrastive learning for general face forgery detection.
\newblock In {\em AAAI}, 2022.

\bibitem{tan2019efficientnet}
Mingxing Tan and Quoc~V Le.
\newblock Efficientnet: Rethinking model scaling for convolutional neural networks.
\newblock {\em ICML}, 2019.

\bibitem{tolosana2020deepfakes}
Ruben Tolosana, Ruben Vera-Rodriguez, Julian Fierrez, Aythami Morales, and Javier Ortega-Garcia.
\newblock Deepfakes and beyond: A survey of face manipulation and fake detection.
\newblock {\em arXiv preprint arXiv:2001.00179}, 2020.

\bibitem{vaswani2017attention}
Ashish Vaswani, Noam Shazeer, Niki Parmar, Jakob Uszkoreit, Llion Jones, Aidan~N Gomez, {\L}ukasz Kaiser, and Illia Polosukhin.
\newblock Attention is all you need.
\newblock {\em Advances in neural information processing systems}, 30, 2017.

\bibitem{wang2023dynamic}
Yuan Wang, Kun Yu, Chen Chen, Xiyuan Hu, and Silong Peng.
\newblock Dynamic graph learning with content-guided spatial-frequency relation reasoning for deepfake detection.
\newblock In {\em Proceedings of the IEEE/CVF Conference on Computer Vision and Pattern Recognition}, pages 7278--7287, 2023.

\bibitem{woo2018cbam}
Sanghyun Woo, Jongchan Park, Joon-Young Lee, and In~So~Kweon.
\newblock Cbam: Convolutional block attention module.
\newblock In {\em ECCV}, pages 3--19, 2018.

\bibitem{xu2023tall}
Yuting Xu, Jian Liang, Gengyun Jia, Ziming Yang, Yanhao Zhang, and Ran He.
\newblock Tall: Thumbnail layout for deepfake video detection.
\newblock In {\em Proceedings of the IEEE/CVF International Conference on Computer Vision}, pages 22658--22668, 2023.

\bibitem{yan2023ucf}
Zhiyuan Yan, Yong Zhang, Yanbo Fan, and Baoyuan Wu.
\newblock Ucf: Uncovering common features for generalizable deepfake detection.
\newblock {\em arXiv preprint arXiv:2304.13949}, 2023.

\bibitem{zhang2023adding}
Lvmin Zhang, Anyi Rao, and Maneesh Agrawala.
\newblock Adding conditional control to text-to-image diffusion models.
\newblock In {\em Proceedings of the IEEE/CVF International Conference on Computer Vision}, pages 3836--3847, 2023.

\bibitem{zhao2021multi}
Hanqing Zhao, Wenbo Zhou, Dongdong Chen, Tianyi Wei, Weiming Zhang, and Nenghai Yu.
\newblock Multi-attentional deepfake detection.
\newblock {\em CVPR}, 2021.

\bibitem{zhao2021learning}
Tianchen Zhao, Xiang Xu, Mingze Xu, Hui Ding, Yuanjun Xiong, and Wei Xia.
\newblock Learning self-consistency for deepfake detection.
\newblock In {\em CVPR}, pages 15023--15033, 2021.

\bibitem{zhao2023diffswap}
Wenliang Zhao, Yongming Rao, Weikang Shi, Zuyan Liu, Jie Zhou, and Jiwen Lu.
\newblock Diffswap: High-fidelity and controllable face swapping via 3d-aware masked diffusion.
\newblock In {\em Proceedings of the IEEE/CVF Conference on Computer Vision and Pattern Recognition}, pages 8568--8577, 2023.

\bibitem{zhuang2022uia}
Wanyi Zhuang, Qi~Chu, Zhentao Tan, Qiankun Liu, Haojie Yuan, Changtao Miao, Zixiang Luo, and Nenghai Yu.
\newblock Uia-vit: Unsupervised inconsistency-aware method based on vision transformer for face forgery detection.
\newblock {\em ECCV}, 2022.

\bibitem{zi2020wilddeepfake}
Bojia Zi, Minghao Chang, Jingjing Chen, Xingjun Ma, and Yu-Gang Jiang.
\newblock Wilddeepfake: A challenging real-world dataset for deepfake detection.
\newblock In {\em ACM MM}, pages 2382--2390, 2020.

\end{thebibliography}
}

\newpage
\appendix
\renewcommand{\thesection}{\Alph{section}} 

\section{Appendix}

\subsection{Cross-manipulation evaluation.}

\noindent\textbf{Cross-manipulation evaluation.}
To further validate the generalization ability across different manipulation methods, we conduct a cross-manipulation evaluation. We train models on a single manipulation method within the high-quality FF++ dataset and test them on all four methods. Using EfficientNet-B4 (En-B4) as the backbone, we compare our approach with the MAT method, which employs attention mechanisms to enhance the generalization ability of En-B4.
Table~\ref{table:2} shows that DiffusionFake improves generalization performance across all manipulation methods. Notably, when trained on the FaceSwap method and tested on the Deepfake method, our approach outperforms the original En-B4 by 6\%. Moreover, compared to the MAT method, DiffusionFake achieves a 4\% improvement in generalization when trained on NeuralTextures and tested on FaceSwap.
These results demonstrate the effectiveness of DiffusionFake in learning generalizable features that can be applied to unseen manipulation methods. 

We also evaluate the multi-source generalization performance by training on three forgery methods and testing on the unknown method. Additionally, we assess the performance under low-quality (LQ) training conditions. As reported in Table~\ref{table:3}, DiffusionFake achieves state-of-the-art results across all protocols and quality levels. Specifically, integrating our method with En-B4 improves generalization by approximately 8\% compared to the backbone alone. Even under low-quality training conditions, DiffusionFake maintains a 7\% performance gain, demonstrating the robustness and generalization capability of our proposed framework.


\begin{table}[h!]
    \begin{minipage}[t]{0.48\textwidth}
        \centering
        \caption{Cross-manipulation evaluation in terms of AUC. Diagonal results indicate the intra-domain performance.}
        \label{table:2}
        \renewcommand\arraystretch{1.2}
        \resizebox{\columnwidth}{!}{
        \begin{tabular}{c|c|cccc}
        \hline
        Train   & Method       & DF    & F2F   & FS    & NT    \\
        \hline
        \multirow{4}{*}{DF}  
        & EN-b4 & \textbf{\textit{99.97}} & 76.32 & 46.24  & 72.72 \\
        & MAT          & \textit{99.91} & 78.23 & 40.61 & 71.08 \\
        & Ours    & \textit{99.82} & \textbf{78.46} & \textbf{52.29} & \textbf{74.43} \\
        \hline
        \multirow{3}{*}{F2F} 
        & EN-b4 & 84.52 & \textit{99.20}  & 58.14 & 63.71 \\
        & MAT          & 86.15&	\textit{99.13}&	60.14&	64.59 \\
        & Ours    & \textbf{88.92} & \textit{\textbf{99.36}} & \textbf{63.19} & \textbf{68.55} \\
        \hline
        \multirow{3}{*}{FS}  
        & EN-b4 & 69.25 & 67.69 & \textbf{\textit{99.89}} & 48.61 \\
        & MAT          & 64.13	&66.39&\textit{99.67}	&	50.10 \\
        & Ours    & \textbf{75.28} & \textbf{70.91} & \textit{99.12} & \textbf{52.17} \\
        \hline
        \multirow{3}{*}{NT}  
        & EN-b4 & 85.99 & 48.86 & 73.05 & \textit{98.25} \\
        & MAT          &  87.23	&48.22&	75.33&	\textit{98.66}  \\
        & Ours    & \textbf{89.54} & \textbf{51.71} & \textbf{79.15} & \textit{\textbf{98.71}} \\
        \hline
                             
        \end{tabular}
        }

        \end{minipage}
        \begin{minipage}[t]{0.48\textwidth}
        \centering
        \caption{Multi-source manipulation evaluation in terms of ACC, which follows~\cite{sun2021domain}. H means high-quality image (c23) in FFpp, while L represents low-quality (c40).}
        \label{table:3}
        \renewcommand\arraystretch{1.2}
        \resizebox{\columnwidth}{!}{
        \begin{tabular}{c|c|c|c|c}
        \hline
        {Method}& {DF (H)} & {DF (L)} & {F2F(H)} & {F2F(L)} \\
        \hline
        Xception&78.25&68.12&61.53&59.58\\
        EN-B4& 82.40& 67.60& 63.32& 61.41\\
        VIT-B&81.15&73.38&62.19&61.93\\
        Multi-task& 70.30& 66.76& 58.74& 56.50\\
        MLDG& 84.21& 67.15& 63.46& 58.12\\
        LTW& 85.60& 69.15& 65.60& 65.70\\
        DCL& 87.70& 75.90& 68.40& 67.85\\
        RECCE&86.69&75.89&62.71&68.02\\
        MAT&84.40&73.71&66.28&66.39\\
        UCF&86.70&74.59&67.87&67.33\\
    

    \hline   
    En-b4+Ours& \textbf{88.17}& 75.13& \textbf{70.17}& \textbf{71.25}\\
    VIT-b+Ours& 87.23& \textbf{77.33}& 68.93& 68.75\\

    \hline

    \end{tabular}
        }

        \end{minipage}
    \end{table}

\subsection{Visualizations of CAM result.}

\begin{figure*}[t!]
    \begin{center}
    \includegraphics[width=1.0\textwidth]{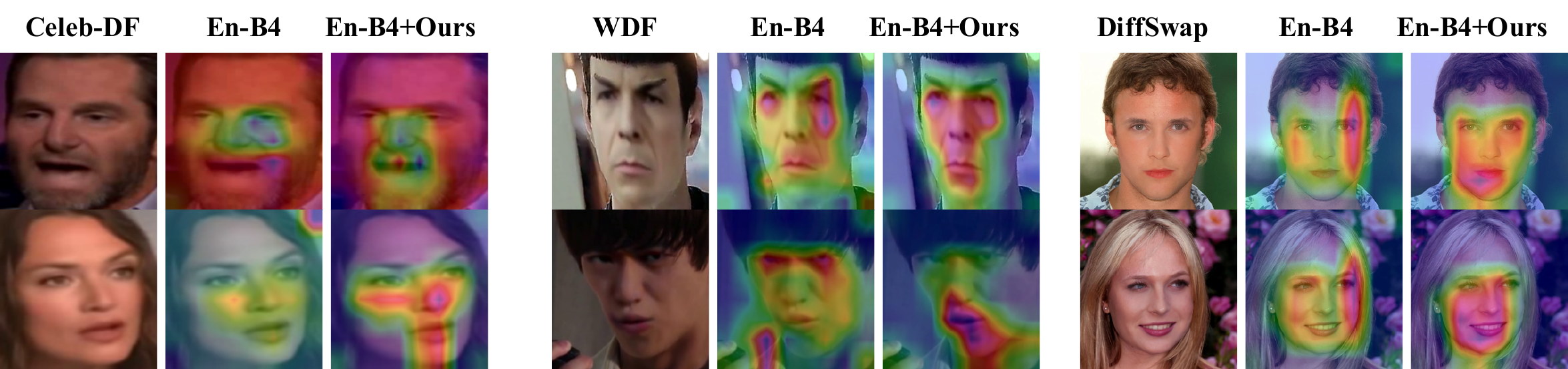}
    \end{center}
       \caption{CAM maps of the baseline model (EN-b4) and En-b4 trained with DiffusionFake method on three unseen datasets: Celeb-DF, WDF (Wild-Deepfake), and DiffSwap.
       }
    \label{fig:vis_cam}
\end{figure*}

To further illustrate the ability of our method to accurately focus on relevant locations in generalized images, we visualize the Class Activation Mapping (CAM) results of both our approach and the vanilla encoder across different datasets. As shown in Figure \ref{fig:vis_cam}, the conventional EfficientNet-B4 (En-B4) encoder often fails to highlight key areas, such as the blurred mouth region in Celeb-DF images. This limitation can reduce the effectiveness of forgery detection. In contrast, our method demonstrates a broader focus during training, targeting significantly larger regions that may include latent forgery areas. This comprehensive attention to detail contributes to enhancing the generalization performance of the detection model. By effectively identifying and concentrating on these critical regions, our method provides a more robust defense against sophisticated forgery techniques, ultimately leading to more accurate and reliable detection outcomes across diverse datasets.

\subsection{Visualizations of weight module.}
\begin{figure*}[t!]
    \begin{center}
    \includegraphics[width=1.0\textwidth]{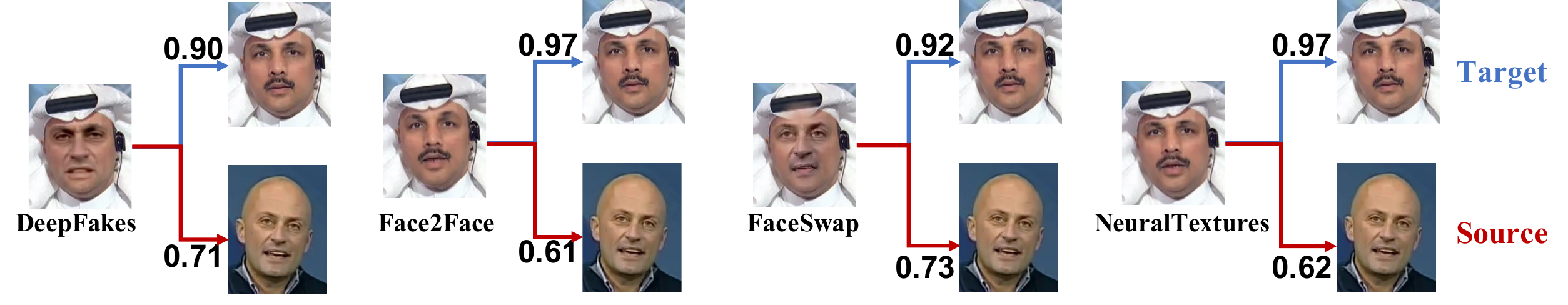}
    \end{center}
       \caption{Visualization of weights for different attack types. The blue lines connect the target weights, while the red lines connect the source weights.
       }
    \label{fig:vis_weight}
\end{figure*}

Figure \ref{fig:vis_weight} presents the source and target scores computed by our Weight Module for four different attack types. It is evident that the target scores are generally higher than the source scores, indicating that reconstructing the target information contributes more significantly to the overall reconstruction process, while the reconstruction of the source image relies more heavily on the pre-trained knowledge. Moreover, the scores vary across different attack types. For samples that are more similar to the target, such as NeuralTextures and Face2Face, the corresponding target scores are higher (greater than 0.95) due to the high proportion of target features they contain, while the source scores are lower due to the limited presence of source features. On the other hand, for Deepfakes and FaceSwap, which involve replacing the source's facial region onto the target, the proportion of source information is relatively higher, resulting in slightly elevated source scores compared to other attack types.

\subsection{Evaluation Metric}
We use two common metrics to evaluate the performance of our forgery detection method: the Area Under the Receiver Operating Characteristic Curve (AUC) and the Equal Error Rate (EER).

The AUC is a widely adopted metric that measures the overall performance of a binary classifier across all possible decision thresholds. It represents the probability that a randomly chosen positive instance (i.e., a forged image) will be ranked higher than a randomly chosen negative instance (i.e., a real image).
The EER is another commonly used metric that represents the point on the ROC curve where the False Positive Rate (FPR) and the False Negative Rate (FNR) are equal.

In summary, we use AUC and EER as our primary evaluation metrics, where a higher AUC and a lower EER indicate better forgery detection performance. These metrics provide a comprehensive assessment of the classifier's ability to distinguish between forged and genuine images across various decision thresholds.

\subsection{Limitations and Broader Impacts.}
\label{sec:limitation}
\textbf{Limitation: }
The framework relies on paired source and target images for training, which may not always be feasible in real-world scenarios. We aim to integrate self-supervised methods to generate these images in the future. Additionally, the effectiveness of DiffusionFake against more sophisticated forgery techniques, such as those involving multiple source identities or partial manipulations, requires further investigation.

\textbf{Broader Impacts: }Our method could potentially be used as an adversarial discriminator to create more difficult-to-detect images. Future research needs to address how to prevent this misuse.

\end{document}